\UseRawInputEncoding
\pdfoutput=1
\RequirePackage{fix-cm}
\documentclass[smallcondensed]{svjour3}     

\usepackage{multirow}
\usepackage{mathtools}
\usepackage{bm}
\usepackage{algorithm}
\usepackage{algpseudocode}
\usepackage{colortbl}
\usepackage[table]{xcolor}
\usepackage{verbatim}
\smartqed  
\usepackage{graphicx}

\begin{document}

\title{Attention Round for Post-Training Quantization
}


\author{Huabin Diao         \and
              Gongyan Li           \and    
              Shaoyun Xu	 	 \and
              Yuexing Hao        
}


\institute{Huabin Diao \at
              Institute of Microelectronics, Chinese Academy of Sciences,
              Beijing,
              100029, 
              China \\
              University of Chinese Academy of Sciences,
              Beijing,
              100049, 
              China\\
              \email{diaohuabin@ime.ac.cn}           
           \and
           Gongyan Li \at
             Institute of Microelectronics, Chinese Academy of Sciences,
             Beijing,
             100029, 
             China
           \and
			Shaoyun Xu \at
			Institute of Microelectronics, Chinese Academy of Sciences,
			Beijing,
			100029, 
			China
           \and
			Yuexing Hao \at
			Institute of Microelectronics, Chinese Academy of Sciences,
			Beijing,
			100029, 
			China
}

\date{Received: date / Accepted: date}

\maketitle

\begin{abstract}
At present, the quantification methods of neural network models are mainly divided into post-training quantization (PTQ) and quantization aware training (QAT). Post-training quantization only need a small part of the data to complete the quantification process, but the performance of its quantitative model is not as good as the quantization aware training. This paper presents a novel quantification method called Attention Round. This method gives parameters $ w $ the opportunity to be mapped to all possible quantized values, rather than just the two quantized values nearby w in the process of quantization. The probability of being mapped to different quantified values is negatively correlated with the distance between the quantified values and $ w $, and decay with a Gaussian function. In addition, this paper uses the lossy coding length as a measure to assign bit widths to the different layers of the model to solve the problem of mixed precision quantization, which effectively avoids to solve combinatorial optimization problem. This paper also performs quantitative experiments on different models, the results confirm the effectiveness of the proposed method. For ResNet18 and MobileNetV2, the post-training quantization proposed in this paper only require 1,024 training data and 10 minutes to complete the quantization process, which can achieve quantization performance on par with quantization aware training.
\end{abstract}

\section{Introduction}
\label{sec:Introduction}

In recent years, deep neural networks have developed rapidly and achieved remarkable results in many domains, such as computer vision, autonomous driving, natural language processing and speech recognition. However, the implementation of deep neural networks often requires huge computational resources and expensive computational costs, severely limiting their application to resource-limited devices. At present, there are three main ways to solve this problem: model compression, lightweight model architecture design and model quantification. Model compression\cite{dai2019nest,ding2018auto,he2017channel,li2016pruning,guo2016dynamic,hu2016network,luo2017thinet,diao2021implementation} trims the redundant parameters in the trained model to reduce model size and alleviate computational burden. Lightweight model architecture design generally uses neural architecture search\cite{liu2018darts,xie2018snas,xu2019pc,chen2019progressive,tao2021compact} to directly design a small network structure. Model quantification\cite{hubara2021accurate,nagel2020up,yu2021any,yang2021bsq,gupta2015deep,zhang2021diversifying,hubara2020improving,yamamoto2021learnable,nahshan2021loss,lee2021network} refers to mapping the model weights and activation values from 32 bit floating points to the fixed point number of lower bits, thus reducing the inference time and power consumption of model and realizing the acceleration. 
The current popular model quantification algorithms can be divided into two categories:  quantization aware training algorithms and post-training quantization algorithms. Among them, the quantization aware training algorithm\cite{choi2018pact,cai2020zeroq,gong2019differentiable,esser2019learned,wang2019haq} generally require sufficient training on the complete training data set to obtain good quantitative results, which requires a lot of computing resources and a long training time. Adequate training data acquisition will also be a great challenge, which is severely limited in practical application. Compare to the quantization aware training algorithms, post-training quantization algorithms\cite{nagel2019data,choukroun2019low,wang2020towards} has higher training efficiency and lower requirements for training data completeness. This quantization algorithm generally requires only a small amount of training data, and can basically complete the quantization process in a few hours. Although the post-training quantization algorithms has the characteristics of high training efficiency, the quantization model will suffer from serious performance degradation when the quantization precision is low. For example, DFQ\cite{nagel2019data} quantify the ResNet18 to 8-bit, the quantified model accuracy is 69.7\%, but when quantified to 4-bit, the accuracy is only 39\%, suffering a large accuracy loss. This is mainly because the parameter space is not equal to the model space, and the quantization error in directly optimizing the parameter space does not minimize the final task loss. Therefore, to improve the performance of the post-training quantization algorithm, we need to focus on the design of the quantization loss function. Some recent work\cite{nagel2020up,li2021brecq} generally directly analyze the degradation of the model space loss function using a Taylor expansion, the second-order error term indicates that the degradation of the loss function can be approximated using the quantified output error of each layer, i.e $E[L(\boldsymbol{w}+\Delta \boldsymbol{w})]-E[L(\boldsymbol{w})] \approx\|\hat{\boldsymbol{w}} \boldsymbol{x}-\boldsymbol{w} \boldsymbol{x}\|_{F}^{2}$. In addition to the loss function, the quantization function is also an important factor affecting the performance of the quantization algorithm. The quantization function is used to map a floating-point number to a fixed-point value with a specific accuracy, which can be expressed as $q(\cdot): R \rightarrow Q_{b}$. Among these,$ Q_b={q_k,\ 1\le k\le N} $,which represents the set of quantized values and N represents the number of quantized values. This paper only considers uniform quantization because it is hardware-friendly. In the process of uniform quantization, the different quantization values in $  Q_b $ have equal intervals. The current commonly used quantization functions are mainly divided into several classes:\\

\textbf{Nearest Round, Floor Round, and Ceil Round}.  Among them, Nearest Round is the most commonly used quantization function in the current quantization methods, which maps the parameters w to the nearest quantization value, i. e $\widehat{\boldsymbol{w}}=s \cdot \operatorname{clip}\left(\lfloor  \frac{w}{s} \rceil, l, h\right)$. s representing the quantized scale parameter,l,h representing the truncation range respectively. When replaced $ \lfloor  \cdot \rceil $ with  $ \lfloor  \cdot \rfloor $ or $ \lceil  \cdot \rceil $, Ceil Round and Floor Round can be obtained. They map w to the nearest quantification value larger than it and smaller than it, respectively. \\

\textbf{Stochastic Round} refers to mapping $ w$ to its two nearest quantified values in the form of probability, i. e. $\widehat{\boldsymbol{w}}=\left\{\begin{array}{l}s \cdot \operatorname{clip}\left(\left \lceil \frac{w}{s} \rceil, l, h\right), \text { with the probability of } \frac{w}{s}-\left \lfloor \frac{w}{s}\right \rfloor \right. \\ \left.s \cdot \operatorname{clip}\left(\lfloor \frac{w}{s}\right \rfloor, l, h\right), \text { with the probability of }\left \lceil \frac{w}{s} \rceil-\frac{w}{s}\right.\end{array}\right.$ \\

\textbf{AdaRound} is an adaptive quantization method that introduces a trainable variable $ V $ into the quantization function, and then constructs complex rectifier functions $ h(V) $ and constraint terms $ f(V) $ based on them. Map $  w $ to the two nearest quantization values by training, i.e.:
$ \hat{w}=s\cdot clip\left(\left\lfloor\frac{w}{s}\right\rfloor+h\left(V\right),l,h\right) $.
The optimized objective function is $\operatorname{argmin}_{V}\|\boldsymbol{w} x-\widehat{\boldsymbol{w}} x\|_{F}^{2}+\lambda f(\boldsymbol{V})$, $\|\cdot\|_{F}^{2}$ represents the Frobenius norm. $ h\left(V_{i,j}\right)=clip\left(sigmoid\left(V_{i,j}\right) \cdot \left(\xi-\gamma\right)+\gamma,\ 0,1\right) $,$ f\left(V\right)=\sum_{i,j}{1-\left|2h\left(V_{i,j}\right)-1\right|^\beta} $, $ \lambda $,$ \xi $,$ \gamma $ are hyperparameters. In the training, control the value of the value of $ \beta $ and make it gradually moving towards 0 or 1 to achieve the goal of the controlling $w $ mapped to the two quantized values closest to it. AdaRound introduces local adaptation in the quantization, which means it can flexibly be mapped to two nearby quantization values, and thus has better results than the first others. \\

In this work, this paper proposes a completely novel quantization function called Attention Round, which treats quantization as a lossy coding process and treats the quantization of w as adding a random perturbation $  \alpha $ on it. Random perturbation$  \alpha $ follows a Gaussian distribution with a mean of 0 and a variance of $  \tau^2 $. This random perturbation gives the parameters the opportunity to map to all possible quantization values while satisfying the closer the quantization values, the larger the probability, and the probability decay according to a Gaussian distribution with the mean of w and variance of $ \tau^2 $. This is like our common attention mechanism, where weights focus greater attention on the quantization values around them, while retaining less attention on further quantization values. \\

The contribution of this paper mainly has the following two aspects:
\begin{itemize}
	\setlength{\itemsep}{0pt}
	\setlength{\parsep}{0pt}
	\setlength{\parskip}{0pt}
	\item Introduce a novel quantization function, Attention Round. It can achieve mapping the weight parameters to all possible quantization values with different attention weights, so that the quantization optimization space is expanded while also ensuring rapid convergence. At the same time, only a small part of the training data and a brief training time are required to achieve the quantitative calibration, and the quantification efficiency is high.
	\item Use the coding length to assign different quantization precision in each layer. Based on rate distortion theory, this paper measures the coding length of each layer, so as to assign different quantization precision, avoid solving the combinatorial optimization problem, and greatly improve the efficiency of mixed precision quantification.

\end{itemize}

\section{Related works}
\label{sec:Related works}
This paper will demonstrate the quantification research works in two ways. \\

\textbf{Quantization aware training (QAT) }is the combination of parameter training and quantification to compensate for the performance degradation caused by the quantification by the training of parameters. During quantization, the quantization function is responsible for mapping floating-points to the fixed-point number, which results in gradient has zero value. At present, most QAT methods directly use straight-through gradient estimation (STE) for gradient approximation, which brings about a large gradient error. To reduce the gradient error, Gong\cite{gong2019differentiable} use a differentiable tanh function to approach the step-quantization function incrementally. DoReFa\cite{zhou2016dorefa} proposes that tailoring the weight range before quantification can reduce the quantization error, because it can reduce the error caused by outliers. PACT\cite{choi2018pact} investigated the effect of the activation value trimming of the different layers on the quantification performance, and found that the optimal cropping range between the different layers had an interlayer dependence. SAT\cite{jin2019towards} studied the updating process of the gradient during quantification training and further improves the quantification performance by adjusting the scale of the weights. Other works, such as Li\cite{li2019additive}, believes that the non-uniform quantization algorithm can achieve better performance than the uniform quantization algorithm while ensuring the inference efficiency. However, non-uniform quantization algorithms are not hardware-friendly, and most of the current hardware can only support uniform quantization algorithms. Although the QAT method can bring about better quantitative performance, this usually requires large training datasets and takes more than 100 GPU hours to achieve better results.\\

Compared to quantization aware training, \textbf{post-training quantification (PTQ) }requires only a short amount of training data in a very short time. In general, most deep learning models can be quantified to 8 bit without suffering from a significant loss of accuracy. Nagel\cite{nagel2020up} proposes that the model can be quantified to 8 bit without requiring any training data and without suffering from the loss of accuracy. However, when quantifying the model to 4 bit or even lower, most of the parameter-space-based quantization algorithms cannot achieve better performance. Nagel\cite{nagel2019data} proposes a layer-by-layer calibration of the quantization algorithm to achieve better performance at the 4 bit quantization. Some works such as, the Banner\cite{banner2019post}, Choukroun\cite{choukroun2019low} found the best quantification results for each layer by optimizing the cropping range. Lin\cite{lin2016fixed} And Dong\cite{dong2019hawq} set the different quantization bit width by calculating the SQNR or Hessian of each layer’s parameter, and thus the quantization accuracy of the model is improved.\\

\section{Methodology}
\subsection{Preliminaries}
\label{sec:Preliminaries}
This section provides a brief introduction to the relevant quantification basics that are covered in this paper.\\

\textbf{Symbol description}: This paper uses lowercase plus bold symbols and capitalization plus bold symbols to represent the vector and tensor, respectively. For example, $ \textbf{W}^{l} $ represents the parameter tensor of the l-th layer of the model, and $ \textbf{w}^{l} $ represents the corresponding parameter vector that will be expanded by $ \textbf{W}^{l} $. $  \widehat{\textbf{W}^{l}} $represents the parameters that will be quantified by $ \textbf{W}^{l} $. $E[\cdot]$ represents the expected operator and $\|\cdot\|_{F}^{2}$ represents the Frobenius norm operator. \\

In general, large quantization errors result from only using quantization functions to quantify parameters. To narrow the quantization error and improve the performance of the quantization, the quantization is also calibrated using the calibration dataset and the error function. The most common error functions are mainly as follows:\\

\textbf{Parameter space:} It is an intuitive way to directly calculate the errors between $ \hat{w} $ and $ w $, i. e $\min \|\widehat{w}-w\|_{F}^{2}, s . t . \widehat{w} \in Q_{b}$. However, there are studies\cite{nagel2020up,li2021brecq} show that merely optimizing the quantization error of parameter space does not enable the best performance of the final quantization model.\\

\textbf{Task space:} To improve the performance of the quantification model, directly optimizing the objective loss function is a method, i.e. $\min E[L(\widehat{w})], s . t . \widehat{w} \in Q_{b}$. Although direct optimization of the objective loss function brings better performance, this method often requires sufficient training data, training resources, and longer training time, which is more suitable for the QAT scenario. In the PTQ scenario, direct optimization of the objective loss function generally lacks ideal results. Because in this case, we often only have the trained full precision weight $w^{*}=\operatorname{argmin}_{w} E[L(w)] w \in R$ and a small part of the calibration data, and hope to complete the quantification process in a short time.\\

\textbf{The Taylor expansion:} By treating the quantification as adding a perturbation to the weight $ w $, you can use the Taylor expansion to analyze the performance degradation of the quantification process\cite{nagel2020up}, i.e.
$E[L(\boldsymbol{w}+\Delta \boldsymbol{w})]-E[L(\boldsymbol{w})] \approx \Delta \boldsymbol{w}^{\boldsymbol{T}} g^{(\boldsymbol{w})}+\frac{1}{2} \Delta \boldsymbol{w}^{\boldsymbol{T}} H^{(\boldsymbol{w})} \Delta \boldsymbol{w}$.
Among these, $g^{(w)}=E\left[\nabla_{\boldsymbol{w}} L\right], H^{(w)}=E\left[\nabla_{\boldsymbol{W}}^{2} L\right]$ are gradient and Hessian matrix of  $ w $, respectively, 
$ \nabla_{\mathbf{w}} $ are the corresponding weight perturbations. For the pre-trained model, the weights $ \mathbf{w}  $ can be considered to have converged to the minimum points, so the gradient can be considered as 0, so that only the Hessian matrix should be considered. However, optimizing its Hessian matrix is very difficult due to $\mathbf{w}  $  has large dimension. AdaRound\cite{nagel2020up} approximate the above error function based on reasonable assumptions as $ 
E\left\lbrack {L\left( \mathbf{w} + \mathrm{\Delta}\mathbf{w} \right)} \right\rbrack - E\left\lbrack {L\left( \mathbf{w} \right)} \right\rbrack \approx \left\| {\hat{\mathbf{w}}\mathbf{x} - \mathbf{w}\mathbf{x}} \right\|_{F}^{2} $, thus avoiding directly calculate Hessian matrix. In the quantization, the quantification of the whole model can be completed by optimizing the above functions layer by layer. Based on this, this paper also uses the above error function as the objective function of the optimization, but a new quantization function is proposed , referring to the subsequent chapters.

\subsection{Motivation}
\label{sec:Motivation}

Assuming that the parameter to be quantified is $ w $, the set of quantized values is expressed as $ Q_{b} = \left\{ q_{k},~1 \leq k \leq N \right\}$. For the Nearest Round, the parameter $ w $  mapped by the Nearest Round to the nearest quantified value, which can be represented as $ NearestRound(w) = \left. \lfloor w \right.\rceil $. Floor Round and Ceil Round map w to the maximum quantization less than w and the minimum quantization greater than $ w $, respectively. Although using these functions can map to different quantization values, these functions all use a fixed mapping method, in which all of the parameters in the model are mapped in the same direction. Then all local rounding errors are accumulate to the overall quantization error, resulting in poor quantization performance. Adaround improves the above method, which is modified for a specific quantization task, and w will adaptively makes the selection from the two nearest quantization values to reduce the overall quantization error. The proposed method can adaptively select an optimal quantization value from all the quantization values for mapping, thus minimizing the overall quantization error.\\

Based on information theory, this paper treats quantification as a lossy coding process. For a given pre-trained model, its parameters of each layer $ \mathbf{w}^{l} = \left\{ \mathbf{w}_{j}^{l} \right\},~l \in \left\lbrack {1,2,\ldots,L} \right\rbrack,~j \in \left\lbrack 1,2,\ldots,N_{i} \right\rbrack $, where $ L $ represents the number of layers of the model,  $ N_j  $ represent the number of parameters of the l-th layer. $ \mathbf{w}_{j}^{l} $ follows Gaussian-like distribution with a mean of $ \mu_l $ and a variance of  $ \delta_l^2 $ \cite{huang2021rethinking}. The quantization of parameter $ \mathbf{w}_{j}^{l} $ is treated as adding random perturbation $ \alpha $ on it, and this random perturbation follows a Gaussian distribution with a mean of 0 and a variance of $ \tau^2 $, i. e $ \alpha \in N\left( 0,\tau^{2} \right) $. For the simplicity, the layer number index $ l $ is omitted. Specifically,
\begin{equation}
{\hat{\mathbf{w}}}_{j} = \mathbf{w}_{j} + \alpha
\label{eq:w}
\end{equation}

Because $ \alpha $ follows a Gaussian distribution with a mean of 0 and a variance of $ \tau^2 $, so $ {\hat{\mathbf{w}}}_{j} $ follows a Gaussian distribution with a mean of $ \mu $ and a variance of $ \tau^{2} + \delta^{2} $, that is, $ {\hat{\mathbf{w}}}_{j} $ has the opportunity to be mapped to all possible quantification values, rather than only two nearby ones, which expands the quantification optimization space of $ {\hat{\mathbf{w}}}_{j} $ to achieve better quantification performance. In addition, the probability of being mapped to different quantized values is inversely correlated with the distance between $ \mathbf{w}_{j} $ and quantized values, that is, it prefers to be mapped to closer quantized values, but also retains the chance of being mapped to other quantized values. Specifically, the probability of being quantified to $ q_k $ is 
\begin{equation}
 q_k = {\int_{\frac{q_{k - 1} + q_{k}}{2}}^{\frac{q_{k} + q_{k + 1}}{2}}\frac{1}{\sqrt{2\pi\tau}}}exp\left( - \frac{\left( x - \mathbf{w}_{j} \right)^{2}}{2\tau^{2}} \right)dx
\label{eq:qk}
\end{equation}

In this paper, we design a new quantization method, which enables the opportunity of $ \mathbf{w}_{j} $ to be mapped to all the quantization values and obtain the best global optimal solution for a specific task, which can be called an attention-based quantization method. Because instead of being mapped to all possible quantization values with the same probability, but with a higher probability to the nearest quantization value, the lower probability maps to the farther away from it, this is like an attention mechanism that maintains high attention to the surrounding quantification value of $ \mathbf{w}_{j} $ and gains less attention as the distance from $ \mathbf{w}_{j} $ increases. This approach both expands the space for quantization optimization and ensures fast quantification convergence.

\subsection{Attention Round}
\label{sec:Attention Round}

In conclusion, the proposal has reasonable motivation, and the specific implementation method will be given in this paper. Parameter $ \mathbf{w} $ for each layer:

\begin{equation}
\hat{\mathbf{w}} = s \cdot clip\left( {\left. \lfloor {\frac{\mathbf{w}}{s} + \mathbf{\alpha}} \right.\rceil,l,h} \right)
\end{equation}

This paper uses uniform quantization, s represents the quantification interval and $ l,h $ represent the cutoff value of the quantification interval. Among these, $ \mathbf{\alpha} $ is a trainable vector, and using $ N\left( 0,\left( \frac{\tau}{s} \right)^{2} \right) $ to achieve the initialization. The best quantified value can be obtained by training $ \mathbf{\alpha} $. This paper does not limit the range of $ \mathbf{\alpha} $ and can effectively map $ {\hat{\mathbf{w}}} $  to all possible quantification values. The backward propagation update is specially set up to map to different quantification values with different attention.

\begin{equation}
\mathbf{\alpha} = \mathbf{\alpha} - \eta~\frac{\partial L}{\partial\mathbf{\alpha}} \\
\end{equation}

\begin{equation}
\frac{\partial L}{\partial\mathbf{\alpha}} = ~\frac{\partial L}{\partial\mathbf{z}}\frac{\partial\mathbf{z}}{\partial\mathbf{\alpha}} \\
\end{equation}

\begin{equation}
\frac{\partial\mathbf{z}}{\partial\mathbf{\alpha}} = \left\{ \begin{matrix}
{0.5 + 0.5*{{erf}\left( \frac{\mathbf{\alpha}}{\sqrt{2}*\frac{\tau}{s}} \right)},~~if~\frac{\partial L}{\partial\mathbf{z}} > 0} \\
{0.5 - 0.5*{{erf}\left( \frac{\mathbf{\alpha}}{\sqrt{2}*\frac{\tau}{s}} \right)},~~otherwise} \\
\end{matrix} \right. \\
\end{equation}

\begin{equation}
{{erf}(x)} = \frac{2}{\sqrt{\pi}}{\int_{0}^{x}{exp\left( - t^{2} \right)dt}}
\end{equation}

\begin{figure}[htbp]
\begin{center}
\includegraphics[width=1\linewidth]{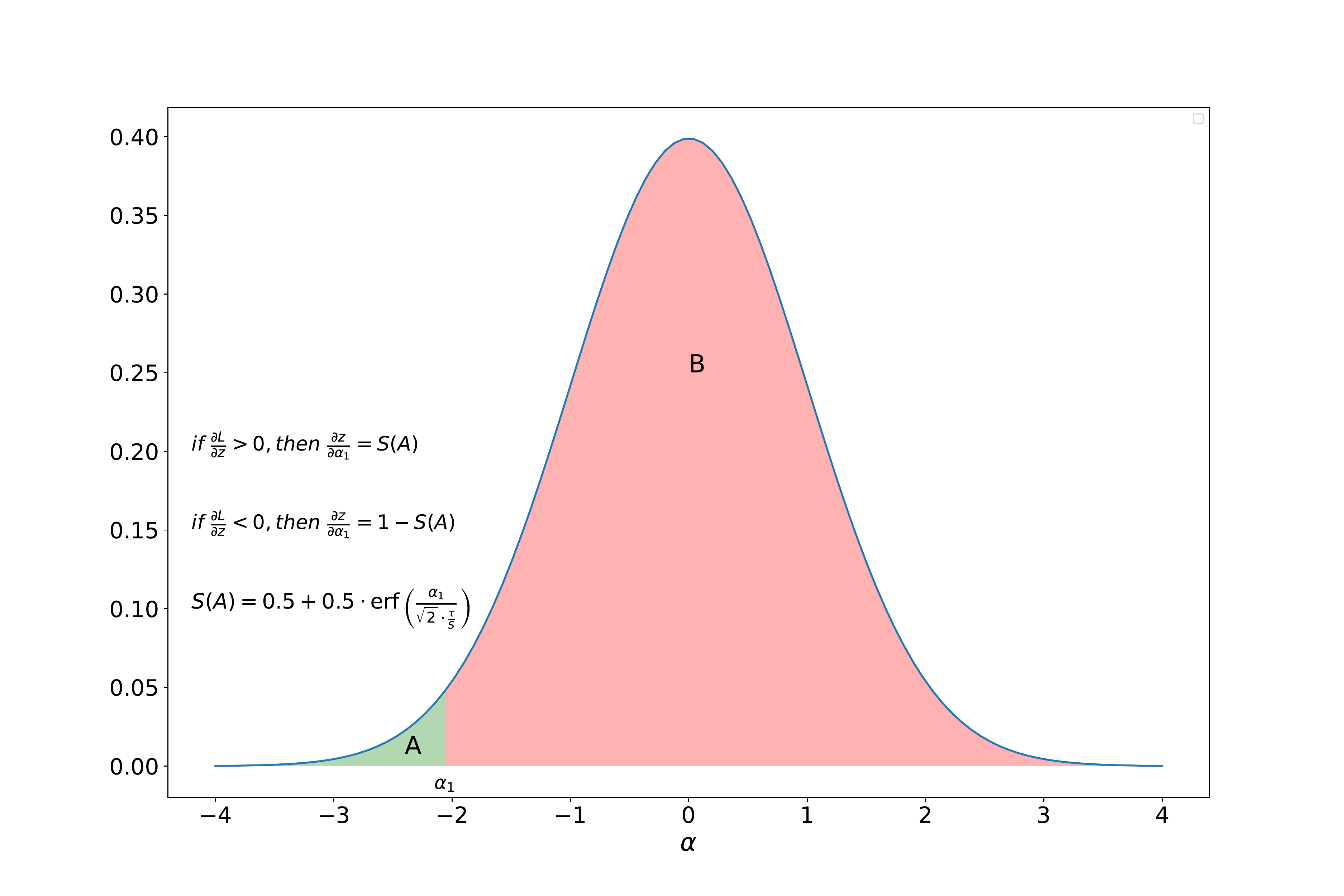}
\end{center}
\caption{The updation of $ \alpha $, $ S(A) $ represents the area of region A}
\label{fig:alpha}
\end{figure}

As shown in Figure \ref{fig:alpha}, we assume $ \mathbf{\alpha} = \mathbf{\alpha}_{1} $,when $ \frac{\partial L}{\partial\mathbf{z}} > 0 $ , the optimization of the loss function is going toward the decrease of $ \mathbf{\alpha}$, $ \left. \frac{\partial z}{\partial\mathbf{\alpha}} \middle| \mathbf{\alpha}_{1} \right. $ is the area of the region A in the figure which denoted as $ S(A) $. Because $ \mathbf{\alpha}_{1} $ is already small at this point, the gradient the gradient corresponding to continuing to decrease $ \mathbf{\alpha}_{1} $ is smaller at this point. When $ 
\frac{\partial L}{\partial\mathbf{z}} < 0 $ , the optimization of the loss function proceeds in the direction that makes $ \mathbf{\alpha}$ increase, at this point $ \left. \frac{\partial z}{\partial\alpha} \middle| \mathbf{\alpha}_{1} \right. $ is $ 1-S(A) $ such that $ \mathbf{\alpha}$ is optimised with a larger gradient towards the neighbourhood of $ \mathbf{w}$. Therefore, when $ \mathbf{\alpha}$ is near $ \mathbf{w}$, it has stronger updating ability, the further the $ \mathbf{\alpha}$ is,  the weaker the updating ability is. That is to say, in the quantitative training process, the nearby quantified values will receive more attention, so that $ \mathbf{w}$ will be mapped to the nearby quantified values with a higher probability, but also retain the attention to other quantified values, so there will be an opportunity to be mapped to other quantified values, but the probability is smaller.

\subsection{Mixed Precision Quantification}
\label{sec:Mixed Precision Quantification}
To reach the limit of post-training quantization, this paper uses a mixed-precision quantization. Given a set of bit widths bits, a bit width is required to assign the parameters of each layer to achieve the best quantization performance of the model. The commonly used quantization algorithms regard the allocation of mixed accuracy as a combinatorial allocation problem, which makes the solution space very large and makes it difficult to solve. \\
From the perspective of rate distortion theory, this paper analyzes the trained parameters of each layer, using the coding length as the measure of the information quantity of each layer, and thus directly assigns the bit width to it. Suppose a given set of vectors $ 
\mathbf{W} = \left( {\mathbf{w}_{1},\mathbf{w}_{2},\ldots,\mathbf{w}_{\mathbf{m}}} \right) \in R^{n \times m} $, this set of vectors should be encoded lossy, and the error after coding satisfy not exceed $ \epsilon $, i.e $ 
E\left\lbrack \left\| {\mathbf{w}_{\mathbf{i}} - {\hat{\mathbf{w}}}_{\mathbf{i}}} \right\|^{2} \right\rbrack \leq \epsilon^{2} $, $ {\hat{\mathbf{w}}}_{\mathbf{i}} $ representing the encoded vectors. For description simplicity, assuming that this set of vectors have a mean of 0, i.e $ \mathbf{\mu} = \frac{1}{m}{\sum_{i}{w_{i} = 0}} $. For each vector $ \mathbf{w}_{\mathbf{i}} $, we can perturb it in a spherical space in an n-dimensional space with a radius of $ \epsilon $. We can simulate this perturbation by adding a Gaussian noise $ z_i $ with the variance of $ \frac{\epsilon^2}{n} $ to each elements of $ \mathbf{w}_{\mathbf{i}} $. We can then get
\begin{equation}
{\hat{\mathbf{w}}}_{\mathbf{i}} = \mathbf{w}_{\mathbf{i}} + \mathbf{z}_{\mathbf{i}}, with \left. \mathbf{z}_{\mathbf{i}}~ \right.\sim~N\left( 0,\frac{\epsilon^{2}}{n}\mathbf{I} \right).
\end{equation}

Then, the covariance matrix of $ {\hat{\mathbf{w}}}_{\mathbf{i}} $ can be expressed as 
\begin{equation}
\hat{\mathbf{\Sigma}} = E\left\lbrack {\frac{1}{m}{\sum\limits_{i = 1}^{m}{{\hat{\mathbf{w}}}_{\mathbf{i}}{\hat{\mathbf{w}}}_{\mathbf{i}}^{\mathbf{T}}}}} \right\rbrack = \frac{\epsilon^{2}}{n}\mathbf{I} + \frac{1}{m}\mathbf{W}\mathbf{W}^{\mathbf{T}}~ \in R^{n \times n}
\end{equation}

The volume of this set of vectors occupying the space is proportional to the square root of the determinant of the covariance matrix, as follows

\begin{equation}
vol\left( \hat{\mathbf{W}} \right) \propto ~\sqrt{\det\left( {\frac{\epsilon^{2}}{n}\mathbf{I} + \frac{1}{m}\mathbf{W}\mathbf{W}^{\mathbf{T}}} \right)}
\end{equation}

Similarly, the volume that random vectors $ {\hat{\mathbf{z}}}_{\mathbf{i}} $ occupy a space is proportional to 
\begin{equation}
vol\left( \mathbf{z} \right) \propto ~\sqrt{det\left( {\frac{\epsilon^{2}}{n}\mathbf{I}} \right)}
\end{equation}

From this, we can obtain that when the error is not exceeded$  \epsilon^2 $, the average number of bits\cite{ma2007segmentation} required for encoding $ \mathbf{W} $ is 
\begin{equation}
L\left( \mathbf{W} \right) = \frac{m + n}{2}{\log_{2}{det\left( \frac{vol\left( \hat{\mathbf{W}} \right)}{vol\left( \mathbf{z} \right)} \right)}} = \frac{1}{2}{\log_{2}{det\left( {I + \frac{n}{m\epsilon^{2}}\mathbf{W}\mathbf{W}^{T}} \right)}}
\label{eq:LW}
\end{equation}

$ L\left( \mathbf{W} \right) $ represents the average number of encoded bits required for a given set of vectors obeying a Gaussian distribution when the error does not exceed $ \epsilon^2 $. Under the same encoding error, the more bits are required, the more information it has. We will assign different quantification accuracy to each layer using $ L\left( \mathbf{W} \right) $ as a measure, with specific steps as described in Algorithm \ref{alg:mixed precision quantization}.

\renewcommand{\algorithmicrequire}{\textbf{Input:}}
\renewcommand{\algorithmicensure}{\textbf{Output:}}
\begin{algorithm}[h]
	\caption{The algorithm for mixed precision quantization}
	\label{alg:mixed precision quantization}
	\begin{algorithmic}[1]
		\Require 
		$ \mathbf{W}_{i},~i \in \left\lbrack {1,2,\ldots,L} \right\rbrack $: Pre-trained parameters for each layer;
		$bit list$: Candidate bit-width collection;
		$ \epsilon^2 $: Encoding error tolerance;
		\Ensure
		 Bit widths for different layers;
		\State Define the encoding length set $ L $;
		\For{$ i $ in $ \left\lbrack {1,2,\ldots,L} \right\rbrack $}
		\State According to Equation (\ref{eq:LW}) compute $L\left( \mathbf{W}_{i} \right)$;
		\State Add $L\left( \mathbf{W}_{i} \right)$ to $ L $;
		\EndFor
		\State Sort the elements in  $ L $ in ascending order;
		\State Set $ len(bits) $ clustering centres to cluster $ L $;
		\State Both the cluster center and the candidate bit widths are sorted in ascending order, assigning the bit widths to the corresponding cluster categories in sequential order;
	\end{algorithmic}
\end{algorithm}

\section{Experiments}
\label{sec:Experiments}
To evaluate the performance of Attention Round, this paper performs quantified experiments using different models with different experimental configurations. This paper first describes the details of the experiment in section \ref{sec:Experimental setup}, then compares Attention Round with other post-training quantization algorithms and quantization aware training in Section \ref{sec:Results on ImageNet} and \ref{sec:Contrast to the Quantization Aware Training}, respectively. Finally, the algorithm is extended to mixed precision quantification and performs experimental analysis.
\subsection{Experimental setup}
\label{sec:Experimental setup}
For all the experiments, the BN layer was parametrically fused with the neighboring convolutional layers, and both the weights and the activation values were uniformly quantified. Prior to applying the Attention Round, the optimal quantification interval s was determined by minimization of $ \left\| {\mathbf{W} - \hat{\mathbf{W}}} \right\|^{2} $, $ \hat{\mathbf{W}} $ represents quantification weights obtained via the rounding-to-nearest . In all quantization experiments, the first and last layers of the model were quantified using 8 bit. For the quantification of the calibration, the 1,024 images of ImageNet was used\cite{deng2009imagenet} to train the parameters $ \mathbf{\alpha} $ of the Attention Round. During the calibration process, the Adam optimizer was used\cite{kingma2014adam} to calibrate each module of the model  individually, with the initial learning rate set at 0.0004 and batch size set to 64, and each module was iteratively trained 2k times. All experiments were performed using the Pytorch\cite{paszke2019pytorch}.

\subsection{Results on ImageNet}
\label{sec:Results on ImageNet}
In this paper, different deep learning model architectures were selected for quantification experiments, including Resnet\cite{he2016deep}，MobilenetV2\cite{sandler2018mobilenetv2}, Regnetx\cite{radosavovic2020designing} and also with the Mnasnet\cite{tan2019mnasnet}. Ordinary convolution operators are included in Resnet, deep separable convolutions in MobilenetV2 and group convolutions in Regnetx. Besides, quantitative experiments on architectures obtained through neural architecture search. First, only the weights were quantified, and the experimental results are shown in Table 1. When the weights were quantified to 6 bit, the quantified model accuracy was comparable to that of the floating-point model. As the width of the quantization bit decreases, the quantization model accuracy gradually decreases. Then, by comparing the method with the most popular post-training quantization algorithms, it is seen that the method achieves better performance than these algorithms when quantifying bit widths of 4bit and 3bit. For example, when the bit width is 4 bit, the restnet18 is quantified with a quantization accuracy of 70.72, and the resnet50 with a quantization accuracy of 76.44. Excellent performance is also achieved on different bit widths and models.

\begin{table}[htbp]
	\centering
	\setlength{\tabcolsep}{1mm}
	\begin{tabular}{ccccccc}
		\hline
		\textbf{Methods}   & \textbf{Bits(W/A)} & \textbf{ResNet-18} & \textbf{ResNet-50} & \textbf{MobileNetV2} & \textbf{RegNet-600MF} & \textbf{MnasNet-2.0} \\ \hline
		Full Prec.         & 32/32              & 71.08              & 77                 & 72.49                & 73.71                 & 76.68                \\ \hline
		DFQ\cite{nagel2019data}        & 8/8                & 69.7               & -                  & 71.2                 & -                     & -                    \\ 
		Ours               & 6/32               & 71.02              & 76.87              & 72.53                & 73.52                 & 76.54                \\ 
		Ours               & 5/32               & 70.95              & 76.77              & 72.35                & 73.45                 & 76.38                \\ \hline
		OMSE\cite{choukroun2019low}       & 4/32               & 67.12              & 74.67              & -                    & -                     & -                    \\ 
		AdaRound\cite{nagel2020up}  & 4/32               & 68.71              & 75.23              & 69.78                & 71.97                 & 74.87                \\ 
		AdaQuant\cite{hubara2020improving}  & 4/32               & 68.82              & 75.22              & 44.78                & -                     & -                    \\ 
		Bit-Split\cite{wang2020towards} & 4/32               & 69.11              & 75.58              & -                    & -                     & -                    \\ 
		Ours               & 4/32               & 70.72              & 76.44              & 71.95                & 72.98                 & 75.86                \\ \hline
		AdaRound\cite{hubara2020improving}  & 3/32               & 68.07              & 73.42              & 64.33                & 67.71                 & 69.33                \\ 
		AdaQuant\cite{nagel2020up}  & 3/32               & 58.12              & 67.61              & 12.56                & -                     & -                    \\ 
		Bit-Split\cite{wang2020towards} & 3/32               & 66.75              & 73.24              & -                    & -                     & -                    \\ 
		Ours               & 3/32               & 69.83              & 75.38              & 69.43                & 70.85                 & 73.5                 \\ \hline
	\end{tabular}
	\caption{The performance of different PTQ algorithms when only quantizing weights.}
    \label{tab:only quantizing weights}
\end{table}

To validate the performance of our quantification algorithm, we quantified both the model weights and the activation values, and the experimental results are presented in Table 2. When the weights and activation values were simultaneously quantified up to 6 bit, the performance of the quantization model obtained using the proposed method was even better than that quantified up to 8 bit using the DFQ method. When the weight and activation value of the model are quantified to 4 bit, the quantification accuracy of resnet18 model is 69.65 and the resnet50 model is 74.89. When the weight and activation value of the model are quantified to 3 bit, the quantification accuracy of resnet18 model is 68.55 and the resnet50 model is 73.86. As can be seen from Tables 1 and 2, the algorithms proposed here all achieved better performance than the other algorithms by using different quantification accuracy for different models.

\begin{table}[htbp]
	\centering
	\setlength{\tabcolsep}{1mm}
	\begin{tabular}{lclllll}
		\hline
		\textbf{Methods}   & \multicolumn{1}{l}{\textbf{Bits(W/A)}} & \textbf{ResNet-18} & \textbf{ResNet-50} & \textbf{MobileNetV2} & \textbf{RegNet-600MF} & \textbf{MnasNet-2.0} \\ \hline
		Full Prec.         & 32/32              & 71.08              & 77                 & 72.49                & 73.71                 & 76.68                \\ \hline
		DFQ\cite{nagel2019data}        & 8/8                                    & 69.7               & -                  & 71.2                 & -                     & -                    \\ 
		Ours               & 6/6                                    & 70.95              & 76.68              & 72.22                & 73.52                 & 76.35                \\ 
		Ours               & 5/5                                    & 70.78              & 76.3               & 70.86                & 73.45                 & 75.39                \\ \hline
		LAPQ\cite{nahshan2021loss}      & 4/4                                    & 60.3               & 70                 & 49.7                 & 57.71                 & 65.32                \\ 
		ACIQ-Mix\cite{banner2019post}   & 4/4                                    & 67                 & 73.8               & -                    & -                     & -                    \\ 
		AdaQuant\cite{nagel2020up}  & 4/4                                    & 67.5               & 73.7               & 34.95                & -                     & -                    \\ 
		Bit-Split\cite{wang2020towards} & 4/4                                    & 67.56              & 73.71              & -                    & -                     & -                    \\ 
		Ours               & 4/4                                    & 69.65              & 74.89              & 65.67                & 70.75                 & 72.23                \\ 
		Ours               & 3/4                                    & 68.55              & 73.86              & 60.96                & 67.68                 & 68.24                \\ \hline
	\end{tabular}
	\caption{The performance of different PTQ algorithms when quantizing weights and activation.}
	\label{tab: quantizing weights and activation}
\end{table}

\subsection{Contrast to the Quantization Aware Training}
\label{sec:Contrast to the Quantization Aware Training}

This paper also compares this method with the quantization aware training algorithm, as shown in Table 3. Compared with the post-training quantification algorithm, the quantization aware training can obtain better quantification accuracy, but the cost is high, so the overall efficiency is low. This paper compares the algorithm with the commonly used quantization aware algorithm in resnet18 and mobilenetV2. It can be seen that the algorithm achieves the same accuracy as the quantization aware training, and it only takes 1,024 data sets to train for 10 minutes to complete the whole quantization process.

\begin{table}[htbp]
	\centering
	\setlength{\tabcolsep}{1mm}
	\begin{tabular}{cccccc}
		\hline
		\textbf{Models}                                                                 & \textbf{Methods} & \textbf{Bits(W/A)} & \textbf{Training Data} & \textbf{GPU hours} & \textbf{Accuracy} \\ \hline
		\multirow{5}{*}{\begin{tabular}[c]{@{}c@{}}ResNet-18\\ FP:71.08\end{tabular}}   & ZEROQ\cite{cai2020zeroq}   & 4/4                & 0                      & 0.008              & 21.2              \\ 
		& PACT\cite{choi2018pact}     & 4/4                & 1.2 M                  & 100                & 69.2              \\ 
		& DSQ\cite{gong2019differentiable}     & 4/4                & 1.2 M                  & 100                & 69.56             \\ 
		& LSQ\cite{esser2019learned}     & 4/4                & 1.2 M                  & 100                & 71.1              \\ 
		& Ours             & 4/4                & 1024                   & 0.15               & 69.65             \\ \hline
		\multirow{5}{*}{\begin{tabular}[c]{@{}c@{}}MobileNetV2\\ FP:72.49\end{tabular}} & PACT\cite{choi2018pact}     & 4/4                & 1.2 M                  & 192                & 61.4              \\ 
		& DSQ\cite{gong2019differentiable}     & 4/4                & 1.2 M                  & 192                & 64.8              \\ 
		& HAQ\cite{wang2019haq}     & Mixed/8            & 1.2 M                  & 384                & 70.9              \\ 
		& Ours             & 4/4                & 1024                   & 0.15               & 65.67             \\ 
		& Ours             & 5/5                & 1024                   & 0.17               & 70.86             \\ \hline
	\end{tabular}
	\caption{The Comparison with QAT algorithms.}
	\label{tab: The Comparison with QAT algorithms}
\end{table}

\subsection{Mixed Precision quantization}
\label{sec:Mixed Precision quantization}
For the pretrained model parameters, this paper uses Algorithm1 to assign an optimal quantized bit width for each layer of the model. The longer the encoding length of each layer parameter represents the more information the layer has, thus assigning a higher quantification accuracy based on it. In This paper, two sets of bit widths were selected for experiments, respectively, [3,4,5,6] and [3,4,5], and then compared with a single-precision quantization, and the experimental results are shown in Table 4. In this paper, Single represents uses single precision quantization, where each layer uses the same bit width. Mixed represents uses mixed precision quantization, namely each layer uses a different bit width. Only the parameters of the convolutional layers involved in the quantization were considered when calculating the model size. As can be seen from the experimental results, the use of mixed precision quantification can improve the accuracy of the quantification model without increasing the model size. For example, when quantifying Resnet18 using the set of bit widths of [3,4,5,6], the accuracy reaches 71.02 when the model size is only 5.5M, which is better than the single-precision quantization when the bit width is 6 bit. Similar results can be found in experiments with other models, which shows that the mixed accuracy allocation algorithm proposed here is effective. Meanwhile, the proposed algorithm can be implemented in a short time with high efficiency.

\begin{table}[htbp]
	\centering
	\setlength{\tabcolsep}{1mm}
	\begin{tabular}{ccccc}
		\hline
		\textbf{Models}                                                                 & \textbf{Single/Mixed}   & \textbf{Bits List} & \textbf{Model Size} & \textbf{Accuracy} \\ \hline
		\multirow{6}{*}{\begin{tabular}[c]{@{}c@{}}ResNet-18\\ FP:71.08\end{tabular}}   & \multirow{2}{*}{Mixed}  & {[}3,4,5,6{]}      & 5.5M                & 71.02             \\ 
		&                         & {[}3,4,5{]}        & 5.21M               & 70.87             \\ 
		& \multirow{4}{*}{Single} & 3                  & 4.19M               & 68.55             \\ 
		&                         & 4                  & 5.81M               & 69.65             \\ 
		&                         & 5                  & 6.98M               & 70.78             \\ 
		&                         & 6                  & 8.38M               & 70.95             \\ \hline
		\multirow{6}{*}{\begin{tabular}[c]{@{}c@{}}ResNet-50\\ FP:77.00\end{tabular}}   & \multirow{2}{*}{Mixed}  & {[}3,4,5,6{]}      & 10.01M              & 76.58             \\
		&                         & {[}3,4,5{]}        & 9.53M               & 75.98             \\ 
		& \multirow{4}{*}{Single} & 3                  & 8.8M                & 73.86             \\ 
		&                         & 4                  & 11.73M              & 74.89             \\ 
		&                         & 5                  & 14.66M              & 76.3              \\ 
		&                         & 6                  & 17.59M              & 76.68             \\ \hline
		\multirow{6}{*}{\begin{tabular}[c]{@{}c@{}}MobileNetV2\\ FP:72.49\end{tabular}} & \multirow{2}{*}{Mixed}  & {[}3,4,5,6{]}      & 0.84M               & 72.15             \\  
		&                         & {[}3,4,5{]}        & 0.835M              & 71.36             \\ 
		& \multirow{4}{*}{Single} & 3                  & 0.82M               & 60.96             \\ 
		&                         & 4                  & 1.1M                & 65.67             \\ 
		&                         & 5                  & 1.37M               & 70.86             \\ 
		&                         & 6                  & 1.64M               & 72.22             \\ \hline
		\multirow{6}{*}{\begin{tabular}[c]{@{}c@{}}RegNet-600MF\\ 73.71\end{tabular}}   & \multirow{2}{*}{Mixed}  & {[}3,4,5,6{]}      & 2.4M                & 73.49             \\ 
		&                         & {[}3,4,5{]}        & 2.28M               & 72.68             \\ 
		& \multirow{4}{*}{Single} & 3                  & 2.12M               & 67.68             \\ 
		&                         & 4                  & 2.82M               & 70.75             \\ 
		&                         & 5                  & 3.53M               & 73.45             \\ 
		&                         & 6                  & 4.23M               & 73.52             \\ \hline
		\multirow{6}{*}{\begin{tabular}[c]{@{}c@{}}MnasNet-2.0\\ 76.68\end{tabular}}    & \multirow{2}{*}{Mixed}  & {[}3,4,5,6{]}      & 4.26M               & 75.87             \\  
		&                         & {[}3,4,5{]}        & 4.24M               & 75.53             \\ 
		& \multirow{4}{*}{Single} & 3                  & 4.17M               & 68.24             \\ 
		&                         & 4                  & 5.56M               & 72.23             \\ 
		&                         & 5                  & 6.95M               & 75.39             \\ 
		&                         & 6                  & 8.34M               & 76.35             \\ \hline
	\end{tabular}
	\caption{The results of mixed precision quantization .}
	\label{tab: The results of mixed precision quantization }
\end{table}

\subsection{Ablation study}
\label{sec:Ablation study}

\subsubsection{Comparison of the different quantization functions}
\label{sec:Comparison of the different quantization functions}
To further illustrate the performance of the Attention Round quantization functions, this paper compares the Attention Round with several common quantization functions, namely, Nearest Round, Floor Round, Ceil Round, Stochastic Round, and AdaRound. All the other experimental conditions are identical except for the quantization function. Taking the ResNet18 model as an example, two sets of experiments were conducted with single-precision quantification. The first group quantified the weight parameters to 4 bit, and the activation value was not quantified. In the second set of experiments, both the weight parameters and the activation value were quantified to 4 bit. As can be seen from the experimental results in Table 5, the quantization result of Floor Round and Ceil Round functions is very poor, because these two quantization functions will bring about large rounding errors, which will be amplified by layer through layer transmission, resulting in very bad final quantification performance. Stochastic Round is a random rounding method that also has poor quantification performance. Nearest Round maps the parameters to the most recent quantization values, greatly reducing the rounding errors, but the quantization performance still not ideal. AdaRound introduces trainable parameters, the parameters can be adaptively mapped to the two quantization values closest to it, thus further reducing the rounding error and improving the quantization performance. Attention Round introduces a quantization mechanism similar to attention, so that parameters can be mapped to any quantization value, and more likely to nearby quantization values, and less likely to distant quantization values. Based on this, it can be considered that Attention Round is an extension of AdaRound, which not only expands the quantitative optimization space, achieves better quantitative performance, but also ensures the rapid convergence of training. Experimental results also show that Attention Round has the best quantization performance among the common quantization functions.

\begin{table}[htbp]
	\centering
	\setlength{\tabcolsep}{1mm}
	\begin{tabular}{cccccccc}
		\hline
		\multirow{2}{*}{\textbf{Models}}                                              & \multirow{2}{*}{\textbf{Bits(W/A)}} & \multirow{2}{*}{\textbf{Nearest Round}} & \multirow{2}{*}{\textbf{Floor Round}} & \multirow{2}{*}{\textbf{Ceil Round}} & \multirow{2}{*}{\textbf{Stochastic Round}} & \multirow{2}{*}{\textbf{AdaRound}} & \multirow{2}{*}{\textbf{Ours}} \\
		&                                     &                                         &                                       &                                      &                                            &                                     &                                \\ \hline
		\multirow{2}{*}{\begin{tabular}[c]{@{}c@{}}ResNet-18\\ FP:71.08\end{tabular}} & 4/32                                & 54.22                                   & 0.09                                  & 0.09                                 & 42.57                                      & 68.71                               & 70.72                          \\ 
		& 4/4                                 & 52.14                                   & 0.08                                  & 0.08                                 & 39.56                                      & 68.55                               & 69.65                          \\ \hline
	\end{tabular}
	\caption{The comparison of different quantization functions.}
	\label{tab: The comparison of different quantization functions }
\end{table}

\subsubsection{Impact of $\tau$ in the Attention Round}
\label{sec:Impact the Attention Round}
\begin{figure}[htbp]
	\begin{center}
		\includegraphics[width=1\linewidth]{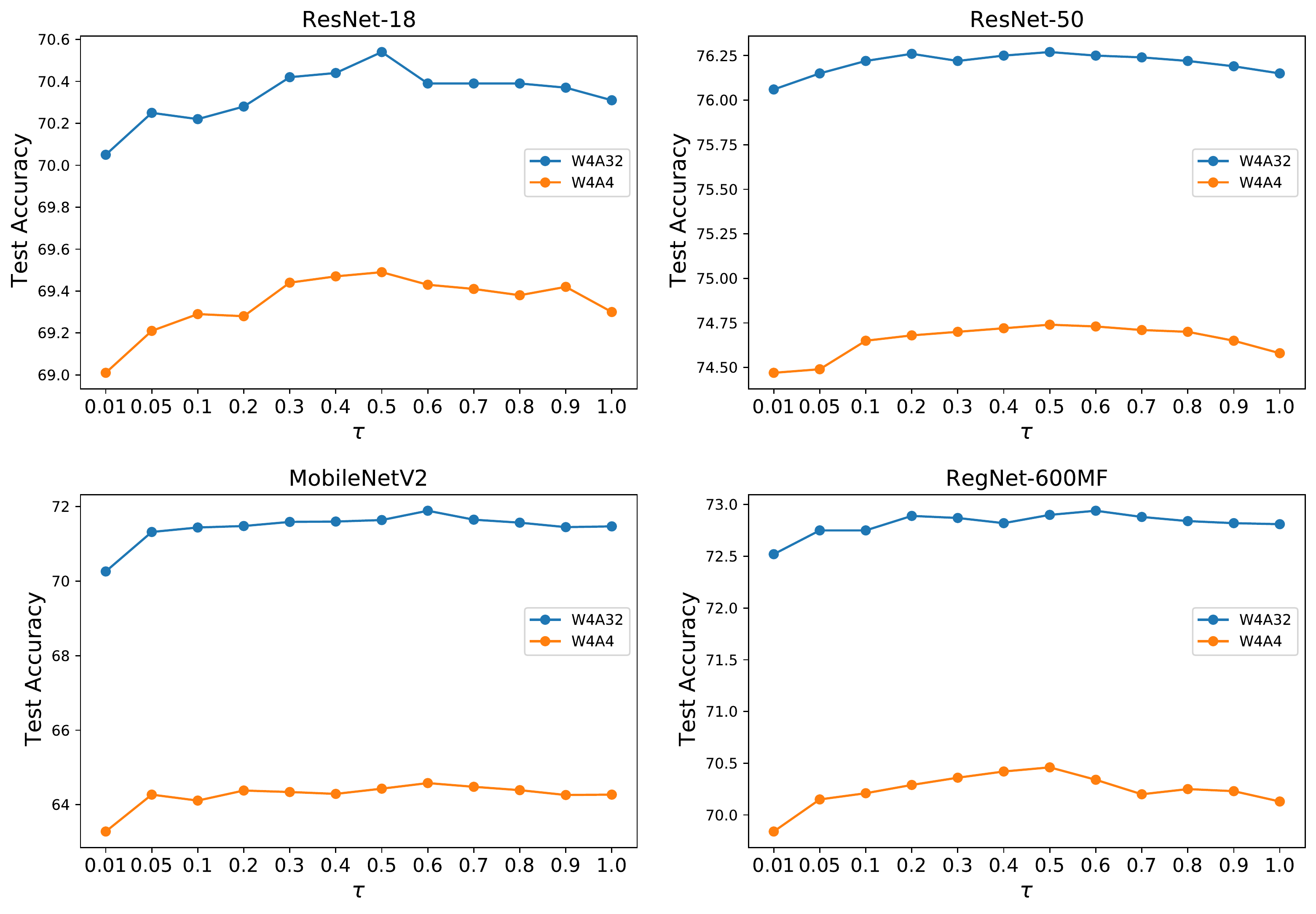}
	\end{center}
	\caption{The effect of $\tau$ on quantization accuracy.}
	\label{fig: The effect on quantization accuracy}
\end{figure}

To evaluate the effect of the unique hyper-parameter $ \tau $ in Attention Round on the accuracy of the model quantification, comparative experiments were conducted for the ResNet18, ResNet50, MobileNetV2, and the RegNet-600MF models. With other experimental settings being the same, different values of $ \tau $ were set to observe their corresponding quantitative model accuracy changes. For each value of $ \tau $, the quantification of the model is divided into two cases. The first case only quantifies the weight parameters, not the activation value. The second case quantifies both the weight parameters and the activation values. The experimental results are shown in Figure \ref{fig: The effect on quantization accuracy}. The quantified model obtained using Attention Round is found to be relatively robust. With $ \tau $ changes from 0 to 1, the accuracy of the quantization model remains essentially stable, fluctuating only over a very small range. Despite the relatively robust performance, better performance can still be achieved with the adjusted values. From the Figure \ref{fig: The effect on quantization accuracy}, when $ \tau $ gradually increases from 0 to 0.5, the quantification accuracy gradually increases, and when it increases to 1, the quantification accuracy gradually decreases, indicating that the best value is around 0.5. Based on this, when using Attention Round, you can safely set the value of $ \tau $ as 0.5.

\subsubsection{Analyze the mixed precision quantification}
\label{sec:Analyze the mixed precision quantification}

In order to analyze the rule of assigning bit widths to different layers of the model in the mixed precision quantization, the set of bit widths is set to [3,4,5,6,7,8], given six different candidate bit widths. In this paper, we obtained the mixed precision assignment results for Resnet18, Resnet50, and MobileNetV2 using Algorithm1, as shown in Figures 2-4. It can be found from the figure that the first and last layers of the model are generally assigned a larger bit width, mainly because the first and last layers of the model often contain rich information and require a larger bit width to achieve better quantification. At the same time, it is also found that the downsampe layer in the model is assigned the minimum bit width, which may also because the downsampe layer itself is only responsible for adjusting the dimension of the feature graph data, which contains less information. In addition, the different convolutional layers within each module are assigned different bit widths, and the intermediate convolution layer is assigned larger bits than the convolutional layers at the two ends, which also reflects the different functions of the convolutional layers at the different depth.

\begin{figure}[htbp]
	\begin{center}
		\includegraphics[width=1\linewidth]{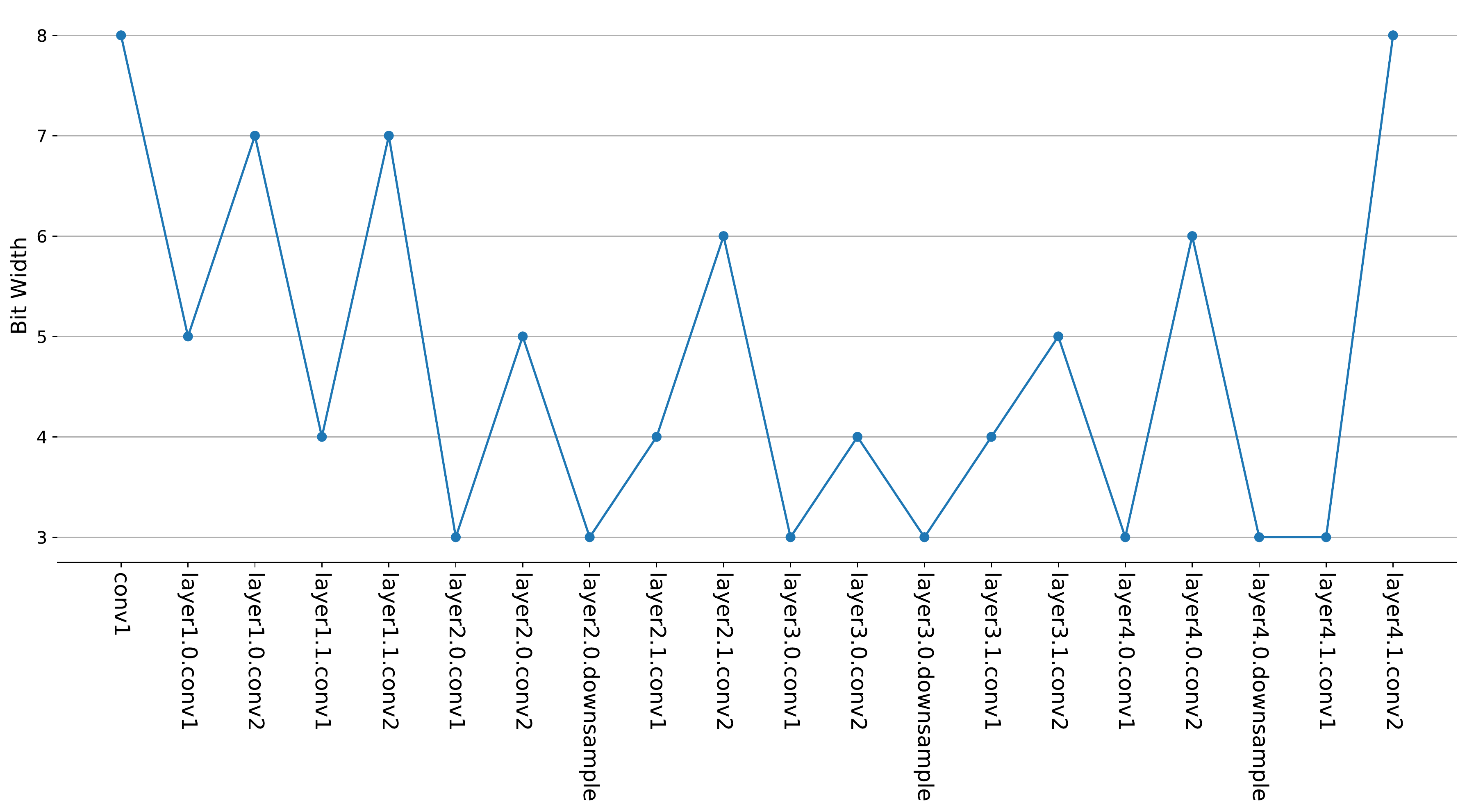}
	\end{center}
	\caption{The bit width of each layer in ResNet-18.}
	\label{fig: The bit width of each layer in ResNet-18}
\end{figure}

\begin{figure}[htbp]
	\begin{center}
		\includegraphics[width=1\linewidth]{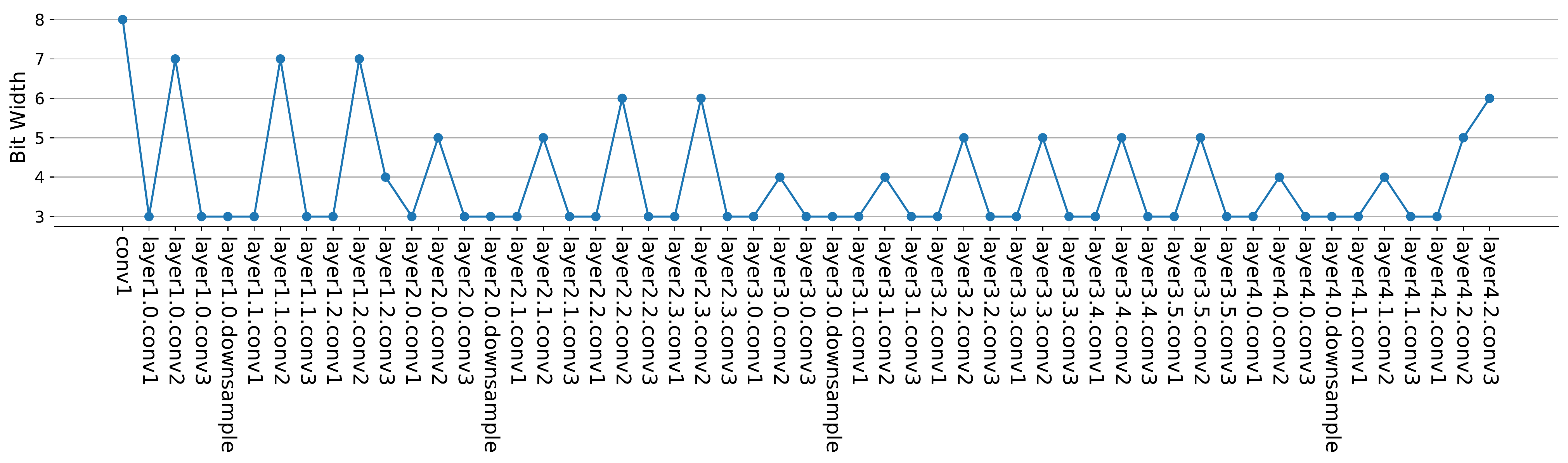}
	\end{center}
	\caption{The bit width of each layer in ResNet-50.}
	\label{fig:The bit width of each layer in ResNet-50.}
\end{figure}

\begin{figure}[htbp]
	\begin{center}
		\includegraphics[width=1\linewidth]{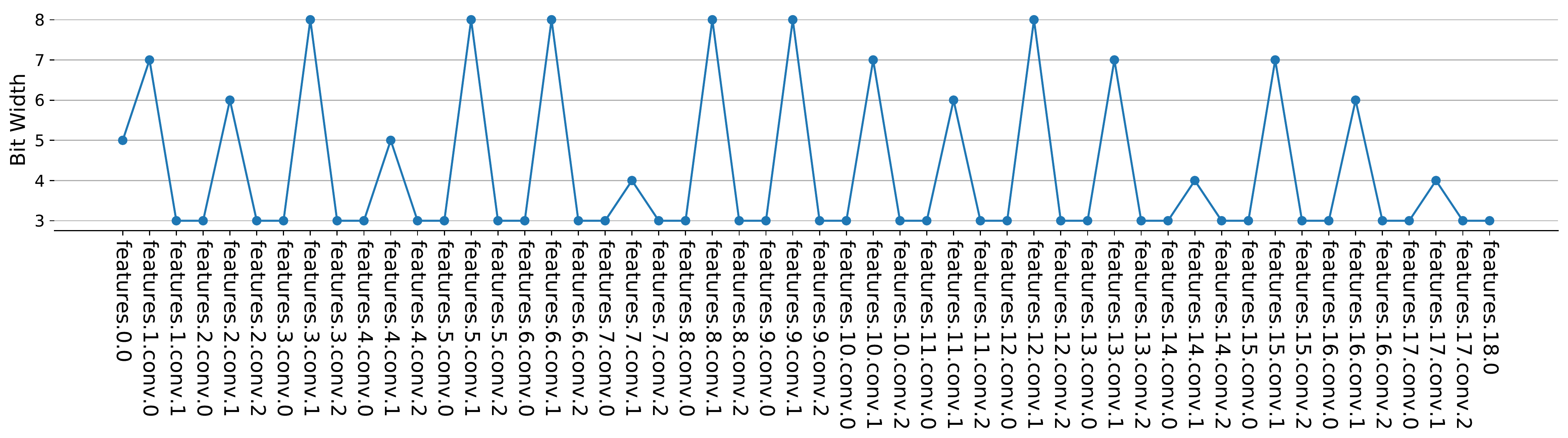}
	\end{center}
	\caption{The bit width of each layer in MobileNet-V2.}
	\label{fig: The bit width of each layer in MobileNet-V2}
\end{figure}

\section{Conclusions}
\label{sec:Conclusions}

This paper based on rate distortion theory, treats quantization as a lossy compression code, thus equivalent analyze the process of quantization of w by adding Gaussian noise$  \alpha $ with the mean of 0 and the variance of $ \tau^2 $ on the parameter $ w $ . Based on the above equivalent analysis, a novel and effective quantification method is designed, called Attention Round. This approach gives it the opportunity to be mapped to all possible quantization values, not just the two quantization values nearby when quantify w.The probability of being mapped to different quantified values is negatively correlated with the distance between the quantified values and w, and is shown to decay with a Gaussian curve with a variance o $ \tau^2 $. Specifically, the probability of $ w $ being quantified to $ q_k $ is $ {\int_{\frac{q_{k - 1} + q_{k}}{2}}^{\frac{q_{k} + q_{k + 1}}{2}}\frac{1}{\sqrt{2\pi\tau}}}exp\left( - \frac{(x - w)^{2}}{2\tau^{2}} \right)dx $. This allows for a large probability of $ w $ being mapped to nearby quantization values, while retaining the possibility of being mapped to other quantization values. On the one hand, this expands the quantitative optimization space for a better quantitative performance. On the other hand, it also ensures the rapid convergence of the training. Also, we use the lossy encoding length as a metric to assign bit widths to the different layers of the model to solve the mixed-precision quantization problem, which avoids solving the combinatorial optimization to quickly determine the optimal quantized bit width for each layer.

 \section*{Conflict of interest}
 The authors declare that they have no conflict of interest.

\bibliographystyle{spphys}       

%
%

\bibliography{Mybib}   
\end{document}